\documentclass[IROS]{RLIpaper}

\newcommand{\e}[1]{ \e{#1}}

\newcommand{\npoints}{n_{\mathcal{P}}}

\usepackage{geometric_algebra}
\usepackage{subfiles}

\author{Cem Bilaloglu, Tobias Löw, and Sylvain Calinon%
\thanks{This work was supported by the State Secretariat for Education, Research and Innovation in Switzerland for participation in the European Commission's Horizon Europe Program through the INTELLIMAN project (\url{https://intelliman-project.eu/}, HORIZON-CL4-Digital-Emerging Grant 101070136) and the SESTOSENSO project (\url{http://sestosenso.eu/}, HORIZON-CL4-Digital-Emerging Grant 101070310).} 
\thanks{
The authors are with the Idiap Research Institute, Martigny, Switzerland and with the Ecole Polytechnique Fédérale de Lausanne (EPFL), Switzerland.
        {\tt\footnotesize cem.bilaloglu@idiap.ch; tobias.loew@idiap.ch; sylvain.calinon@idiap.ch}}%
}

\title{Diffusion-based Virtual Fixtures}

\begin{document}

\maketitle

    \begin{abstract}

        Virtual fixtures assist human operators in teleoperation settings by constraining their actions. This extended abstract introduces a novel virtual fixture formulation \emph{on surfaces} for tactile robotics tasks. Unlike existing methods, our approach constrains the behavior based on the position on the surface and generalizes it over the surface by considering the distance (metric) on the surface. Our method works directly on possibly noisy and partial point clouds collected via a camera. Given a set of regions on the surface together with their desired behaviors, our method diffuses the behaviors across the entire surface by taking into account the surface geometry. We demonstrate our method's ability in two simulated experiments (i) to regulate contact force magnitude or tangential speed based on surface position and (ii) to guide the robot to targets while avoiding restricted regions defined on the surface. All source codes, experimental data, and videos are available as open access at \url{https://sites.google.com/view/diffusion-virtual-fixtures}.

        

    \end{abstract}



\section{INTRODUCTION}
\label{sec:introduction}


For a long time, robotics considered objects in the environment primarily as obstacles and the goal was to avoid contact due to modeling and sensing difficulties. However, the trend has shifted towards embracing contact due to increasing interest in manipulation, tactile robotics, and surface inspection tasks. Consequently, robots physically interact with their surrounding environment that can charecterized by curved surfaces, which can also be soft and fragile (e.g., surgical robotics). However, safety in these tasks remains a major concern during deployment in real-world as they involve forceful interactions. Considering that a significant percentage of recent approaches propose learning-based controllers, and that the majority of shared control and teleoperation tasks depend on the operator's expertise or skills, safety takes a more central role in assistive systems. Therefore, in tactile tasks, there is a need for modules that can be combined with other controllers to enforce high priority constraints.
    
\begin{figure}[tbp]
        \centering
        \includegraphics[width=0.9\linewidth]{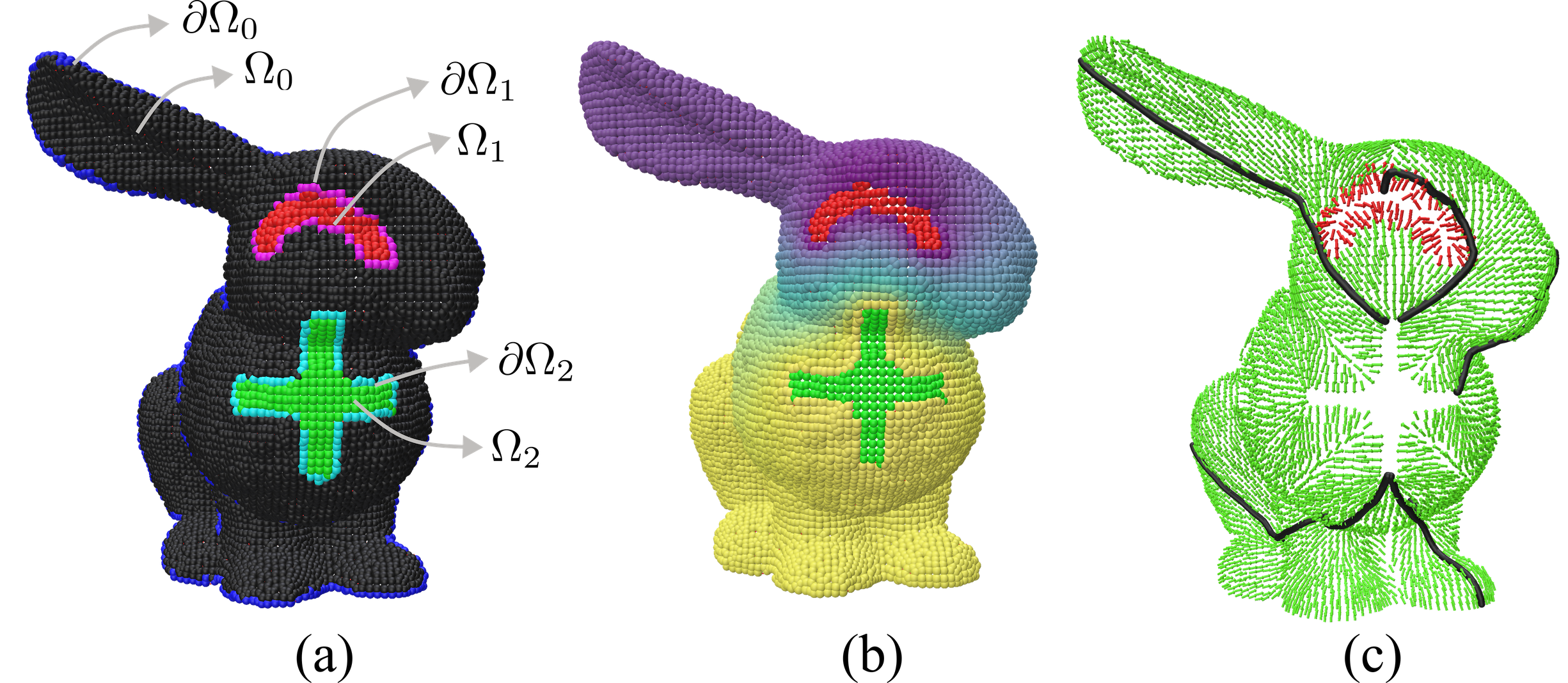}
 
        \caption{Input and example use-cases of our method. Left, the input to our method -- a partial point cloud with segmented regions $\Omega_1$, $\Omega_2$ and the free region $\Omega_0$ with their estimated boundaries $\partial\Omega_1$, $\partial\Omega_2$ and $\partial\Omega_0$. Center, the first experiment showing the virtual fixture constraining the contact force magnitude based on diffusing the contact forces designated on $\Omega_1$ and $\Omega_2$. Right, the second experiment showing the flow field guiding the agents towards target $\Omega_2$ by avoiding restricted regions $\Omega_1$ where the black paths show trajectories with different initial positions.}
               \label{fig:overview}
\end{figure}
Virtual fixtures, introduced by Rosenberg \cite{rosenbergVirtualFixturesPerceptual1993} for teleoperation, are formulations that assist the operator by constraining their actions. Existing methods \cite{selvaggioPassiveVirtualFixtures2018,muhlbauerProbabilisticApproachMultiModal2024,pruksMethodGeneratingRealtime2022a,quinteroRobotProgrammingAugmented2018,quinteroFlexibleVirtualFixture2017}, although operating on surfaces, consider the problem in the robot workspace with the standard Euclidean metric. Accordingly, they do not capture the tangent space and the distance (metric) on the surface. In contrast, various tactile tasks have constraints based on the position on the surface instead of a single guidance path and they change based on the distance on the surface. For instance, to regulate the contact force magnitude or the speed based on the position on the surface, one can annotate the desired values at particular points, as shown in Figure \ref{fig:overview}-a. Then, these values can be diffused by considering the surface metric across the entire surface as in Figure \ref{fig:overview}-b. Alternatively, by specifying only the target and obstacle regions, a smooth flow field on the tangent space can guide agents to the closest target while avoiding the restricted zones and maintaining contact with the surface, as depicted in Figure \ref{fig:overview}-c. For addressing these challenges, we propose a surface virtual fixture method expecting surfaces as possibly noisy and partial point clouds collected in runtime using an off-the-shelf camera attached to the robot. Next, we segment the point cloud into a set of regions with their specified behavior. This segmentation can come from learning-based methods using vision \cite{caronEmergingPropertiesSelfSupervised2021} or geometry \cite{sharpDiffusionNetDiscretizationAgnostic2022a}. Alternatively, one can use virtual or real-world expert annotations \cite{quinteroRobotProgrammingAugmented2018,quinteroFlexibleVirtualFixture2017}, possibly in combination with feature extractors \cite{pruksMethodGeneratingRealtime2022a}. Then, we diffuse the specified behavior across the whole surface to combine them consistently by considering the surface metric.

    



\section{METHOD}
\label{sec:method}

    
    We assume the surfaces that we interact with are two-dimensional Riemannian manifolds $\mathcal{M}$ that we can measure using an RGB-D camera. Accordingly, we can represent them as  \emph{point cloud}s
    \begin{equation}
\Omega := \left\{ (\bm{p}_i, \bm{c}_i) \ \middle| \
\begin{aligned}
&\bm{p}_i \in \mathcal{M}, \ \bm{c}_i \in \{0, \ldots, 255\}^3 \\
&\text{for} \ i = 1, \ldots, \npoints
\end{aligned}
\right\}
\end{equation}
    where $\bm{p}_i$ are the point positions and $\bm{c}_i$ are the RGB color intensities. As explained in the introduction we can find a mapping from $\bm{p}_i$ and $\bm{c}_i$ to a set of specified behaviors $\Phi:=\left\{\phi_1,\ldots, \phi_i,\ldots, \phi_N \right\}$ on these points. We assume the union of points with the same specified behavior $\phi_i$ composes a region $\Omega_i$ on the manifold. We consider all the points that do not have a specified behavior as the free region and denote it with $\Omega_0$. Therefore we segment our manifold $\mathcal{M}$ into $N+1$ disjoint regions. Next, we estimate the boundaries of these regions using the procedure implemented by the \emph{PCL} library and we designate the boundary of the $i$-th region using $\partial \Omega_i$. We provide an example with $N=2$ in Figure \ref{fig:overview}-a.

    Our method aims to generalize the correct behavior to free region $\Omega_0$  given a sparse set of behaviors in $\Omega_i$ considering the metric of the surface. For that purpose, we solve the diffusion equation
    \begin{equation}
            \dot{u}(\bm{x},t) =  \bm{L} u(\bm{x},t) \quad  \text {on} \Omega,
        \label{eq:heat}
    \end{equation}
    where $u(\bm{x},t): \Omega \rightarrow \mathbb{R}$ is the scalar field at time $t$ resulting from diffusing the initial field $u(\bm{x},0)$ and $\bm{L}$ denotes the Laplace-Beltrami operator generalizing the Laplacian $\Delta$ to the Riemannian manifolds. Note that a partial differential equation only specifies the behavior on the interior of the domain, and for the desired unique solution one specifies the behavior at the boundary
    \begin{equation}
            u(\bm{x})=g(\bm{x}) \quad \text { on } \partial \Omega_D, \quad 
            \frac{\partial u(\bm{x})}{\partial n_x} =h(\bm{x}) 
            \quad \text { on } \partial \Omega_N,
            \label{eq:boundaries}
    \end{equation}
    where $\partial \Omega_D$ and $\partial \Omega_N$ correspond to the Dirichlet and Neumann boundaries, respectively. We assume the designated behaviors fix the boundary conditions on the neighboring boundaries of the free region (see cyan and magenta regions in Figure \ref{fig:overview}-a). However, there might be also additional boundaries without specified behavior (e.g., blue regions in Figure \ref{fig:overview}-a). In that case, we assume the boundary to be zero Neumann $h(x) = 0$, meaning no information diffuses through the boundary.

    Note that the solution of the diffusion equation \eqref{eq:heat} depends on the diffusion time $t_D$. After a sufficient time $t_D \gg 0$ the diffusion reaches a steady-state where the time derivative vanishes $\dot{u} (\bm{x},t) = 0$. If we plug this condition into the diffusion equation \eqref{eq:heat}, we recover the steady-state diffusion equation (i.e., Laplace's equation)
    \begin{equation}
             \bm{L} u(\bm{x}) = 0 \quad  \text { on } \Omega.
        \label{eq:laplace}
    \end{equation}
    Laplace's equation \eqref{eq:laplace} minimizes the Dirichlet energy, thus providing the smoothest possible interpolation of the Dirichlet boundary conditions to the interior of the domain. Similarly, the transient diffusion \eqref{eq:heat} provides a locally smooth interpolation, where the radius of the local zone is determined by the diffusion time $t_D$. In the limiting case where $t_D \rightarrow 0$ it is shown that one can recover geodesics \cite{craneGeodesicsHeatNew2013} and interpolate the boundary data based on the geodesic distance. We use the diffusion time as the hyperparameter of our method to select the desired smoothness of the results.

    \section{EXPERIMENTS}
    \label{sec:experiments}
    For the experiments, we used the partial and noisy point cloud of the Stanford bunny, and we assumed an expert marked the regions $\Omega_1$ and $\Omega_2$ as shown in Figure \ref{fig:overview}-a.
    
    As the first experiment, we considered a task where the robot needs to apply contact forces $F_{1}$ and $F_{2}$ in $\Omega_1$ and $\Omega_2$ respectively. Then, we solved Laplace's equation using $F_{1}$ and $F_{2}$ as Dirichlet boundary conditions on the $\partial \Omega_{1_D}$ and $\partial \Omega_{2_D}$ to find the contact forces interpolating the values to $\Omega_0$ as shown in Figure \ref{fig:overview} b.

    For the second task, we considered target-reaching and obstacle-avoidance on the surface. We specified the desired behavior as follows:
    \begin{itemize}
        \item Obstacles $\Omega_1$ push the agents on their interior and do not affect the agents on their exterior. 
        \item Targets $\Omega_2$ do not affect the agents in their interior and pull the agents on their exterior. 

        \item The free region $\Omega_0$ guides the agents toward targets $\Omega_2$ without passing through the obstacles $\Omega_1$
    \end{itemize}
    and satisfy these behaviors using two independent virtual fixtures. First, for obstacle avoidance, we solved the diffusion equation \eqref{eq:heat} in the region $\Omega_1$ where we set the Dirichlet boundary as $\partial\Omega_{1_D} = \bm{1}$ and started from the initial condition $u(\bm{x},0) = \bm{0}\quad\text{in}\quad \Omega_1$. This leads to the gradients in the region $\nabla_{\bm{x}}, \quad \bm{x} \in  \Omega_1 $ point towards the exterior $\nabla_{\bm{x}} =\bm{n}_{\bm{x}}$ where $\bm{n}_{\bm{x}}$ are the unit vectors pointing outward from the region. 
    
    Secondly, for target reaching, we solved the diffusion equation \eqref{eq:heat} in the combined region $\Omega_0 \cup \partial\Omega_2$, where we set the Dirichlet boundary $\partial\Omega_{2_D} = \bm{1}$ and the Neumann boundary $\partial\Omega_{0_N}=\bm{0}$, starting from the initial condition $u(\bm{x},0)= \bm{0} \quad \text{in}\quad \Omega_0$. Accordingly, the gradients at the boundary of the target $\nabla_{\bm{x}}, \quad \bm{x} \in \partial \Omega_2$ point towards the target interior $\nabla_{\bm{x}} = -\bm{n}_{\bm{x}}$. Note that the zero Neumann boundary condition ensures that none of the gradients at the boundary of the free region $\nabla_{\bm{x}}\quad\bm{x}\in\partial\Omega_0$ would point towards the obstacle region or out of the surface. As recommended by \cite{craneGeodesicsHeatNew2013}, we set the time parameter $t_D=h^2$, where $h$ is the mean distance between the neighboring points in the point cloud.

    \section{CONCLUSION \& FUTURE WORK}
    \label{subsec:future_work}
    We presented a novel virtual fixture method on surfaces based on mature tools from the partial differential equations and computer graphics literature for addressing tactile robotics tasks \cite{belkinConstructingLaplaceOperator2009,craneGeodesicsHeatNew2013}. We demonstrated our approach based on diffusing scalar valued functions on surfaces is promising for generalizing the desired behavior defined in some regions to the whole surface, by considering the metric of the surface.
    
    A natural extension of our work is to consider diffusion of vector-valued data \cite{sharpVectorHeatMethod2019,robert-nicoudIntrinsicGaussianVector2024} such that we can use it for diffusing velocities on the tangent space or to diffuse orientations by using the Lie algebra of the unit quaternion group. Alternatively, we can consider bivectors to diffuse the specified dynamical behaviors \cite{ficheraHybridQuadraticProgramming2023,ficheraLearningDynamicalSystems2024a} at particular regions to the whole domain. Another promising direction is to extend our approach from surfaces to other domains such as the workspace $\mathbb{R}^3$ or the joint space of the robot endowed with a non-Euclidean metric, using Monte Carlo methods for solving the diffusion \cite{sawhneyGridfreeMonteCarlo2022, sabelfeldRandomWalkRectangles2019, muchachoWalkSpheresPDEbased2024}.

    

\printbibliography
\end{document}